\documentclass[letterpaper, 10 pt, journal, twoside]{cls/IEEEtran}
\IEEEoverridecommandlockouts		
\usepackage{amsmath}
\usepackage{amssymb}
\usepackage{pifont}

\usepackage{subcaption}
\usepackage{tabularx}
\usepackage{graphicx}
\usepackage{bm}
\usepackage{color}
\usepackage[dvipsnames]{xcolor}
\usepackage{float}
\usepackage{units}
\usepackage{comment}
\usepackage{cancel}
\usepackage[breaklinks=true, colorlinks, bookmarks=true]{hyperref}
\usepackage{cite}

\usepackage{multirow}
\usepackage{diagbox}
\usepackage{balance}
\usepackage{caption}
\usepackage{subcaption}
\usepackage{booktabs,makecell}
\usepackage{ulem} 
\usepackage{xcolor} 

\newif\ifshowredst
\showredstfalse 

\newcommand{\rebut}[1]{{\color{black}#1}}

\newcommand{\redst}[1]{\ifshowredst \textcolor{red}{\sout{#1}} \fi}

\newtheorem{problem}{Problem}

\usepackage{multirow}
\newcommand{\bff}{\mbox{\boldmath $f$}}

{\end{tabbing} \end{mdframed} \vspace{5px}}

\newcommand{\BM}{\begin{bmatrix}}
\newcommand{\EM}{\end{bmatrix}}

\newcommand{\beq}{\begin{equation}}
\newcommand{\eeq}{\end{equation} }

\newcommand{\calA}{{\cal A}}
\newcommand{\calB}{{\cal B}}
\newcommand{\calC}{{\cal C}}
\newcommand{\calD}{{\cal D}}

\newcommand{\calF}{{\cal F}}

\newcommand{\calL}{{\cal L}}

\newcommand{\calU}{{\cal U}}

\newcommand{\frakq}{{\mathfrak{q}}}

\newcommand{\bfc}{\mathbf{c}}

\newcommand{\bfe}{\mathbf{e}}

\newcommand{\bfh}{\mathbf{h}}

\newcommand{\bfp}{\mathbf{p}}
\newcommand{\bfq}{\mathbf{q}}
\newcommand{\bfr}{\mathbf{r}}

\newcommand{\bfu}{\mathbf{u}}
\newcommand{\bfv}{\mathbf{v}}

\newcommand{\bfx}{\mathbf{x}}

\newcommand{\bfalpha}{\boldsymbol{\alpha}}
\newcommand{\bfbeta}{\boldsymbol{\beta}}

\newcommand{\bfzeta}{\boldsymbol{\zeta}}

\newcommand{\bftheta}{\boldsymbol{\theta}}

\newcommand{\bfpi}{\boldsymbol{\pi}}

\newcommand{\bftau}{\boldsymbol{\tau}}

\newcommand{\bfA}{\mathbf{A}}
\newcommand{\bfB}{\mathbf{B}}
\newcommand{\bfC}{\mathbf{C}}
\newcommand{\bfD}{\mathbf{D}}

\newcommand{\bfG}{\mathbf{G}}
\newcommand{\bfH}{\mathbf{H}}
\newcommand{\bfI}{\mathbf{I}}
\newcommand{\bfK}{\mathbf{K}}

\newcommand{\bfP}{\mathbf{P}}
\newcommand{\bfQ}{\mathbf{Q}}
\newcommand{\bfR}{\mathbf{R}}
\newcommand{\bfS}{\mathbf{S}}

\newcommand{\bfU}{\mathbf{U}}

\newcommand{\bfX}{\mathbf{X}}

\newcommand{\bbR}{\mathbb{R}}

\begin{document} 

\title{\Large \bf 			
\rebut{Variable-Frequency} Model Learning and Predictive Control for \\ \rebut{Jumping} Maneuvers on Legged Robots
}

\author{Chuong Nguyen$^1$, Abdullah
Altawaitan$^2$, Thai Duong$^2$, Nikolay Atanasov$^2$, and Quan Nguyen$^1$
\thanks{$^1$The authors are with the Department of Aerospace and Mechanical Engineering, University of Southern California, Los Angeles, CA 90007, USA, e-mails: {\tt\small \{vanchuong.nguyen,\allowbreak quann\}@usc.edu}.}%
\thanks{$^2$The authors are with the Department of Electrical and Computer Engineering, University of California San Diego, La Jolla, CA 92093, USA, e-mails: {\tt\small \{aaltawaitan,\allowbreak tduong,\allowbreak natanasov\}@ucsd.edu}. A. Altawaitan is also affiliated with Kuwait University as a holder of a scholarship.}
}%

												
\maketitle

\begin{abstract}
Achieving both target accuracy and robustness in dynamic maneuvers with long flight phases, such as high or long jumps, has been a significant challenge for legged robots. To address this challenge, we propose a novel learning-based control approach consisting of model learning and model predictive control (MPC) utilizing a \rebut{variable-frequency}\redst{adaptive-frequency} scheme. Compared to existing MPC techniques, we learn a model directly from experiments, accounting not only for leg dynamics but also for modeling errors and unknown dynamics mismatch in hardware and during contact. Additionally, learning the model with \rebut{variable-frequency} \redst{adaptive-frequency} allows us to cover the entire flight phase and final jumping target, enhancing the prediction accuracy of the jumping trajectory. Using the learned model, we also design \rebut{variable-frequency} \redst{adaptive-frequency MPC} to effectively leverage different jumping phases and track the target accurately. 
\rebut{In a total of $92$ jumps on Unitree A1 robot hardware, we verify that our approach outperforms other MPCs using fixed-frequency or nominal model, reducing the jumping distance error $2-8$ times.}
\redst{In hardware experiments with a Unitree A1, we demonstrate that our approach outperforms baseline MPC using a nominal model, reducing the jumping distance error up to 8 times.} 
We also achieve jumping distance errors of less than $3\%$ during continuous jumping on uneven terrain with randomly-placed perturbations of random heights (up to $4$ cm or $27\%$ the robot's standing height). Our approach obtains distance errors of $1-2$ cm on $34$ single and continuous jumps with different jumping targets and model uncertainties. \rebut{Code is available at \url{https://github.com/DRCL-USC/Learning_MPC_Jumping}}.
\end{abstract}
\section{Introduction}
\redst{Aggressive maneuvers with legged robots, such as jumping \cite{QuannICRA19,RL_jump_bipedal}, back-flipping \cite{unitreeH1}, and barrel rolls \cite{chuongjump3D}, have received significant attention recently, and have been successfully demonstrated using trajectory optimization \cite{QuannICRA19,chuongjump3D,matthew_mit2021_1,ChignoliICRA2021}, model-based control \cite{YanranDingTRO,continuous_jump_bipedal,GabrielICRA2021,park2017high, ZhitaoIROS22,fullbody_MPC}, and learning-based control \cite{zhuang2023robot, RL_jump_bipedal,yang2023cajun,vassil_drl,jumpingrl,lokesh_ogmp}.} 
\rebut{Aggressive jumping maneuvers with legged robots have received significant attention recently, and have been successfully demonstrated using trajectory optimization \cite{QuannICRA19,chuongjump3D,matthew_mit2021_1,ChignoliICRA2021}, model-based control \cite{YanranDingTRO,continuous_jump_bipedal,GabrielICRA2021,park2017high, ZhitaoIROS22,fullbody_MPC}, and learning-based control \cite{zhuang2023robot, RL_jump_bipedal,yang2023cajun,vassil_drl,jumpingrl,lokesh_ogmp}}.
Unlike walking or running, aggressive motions are particularly challenging due to \textit{(1)} the extreme underactuation in the mid-air phase (the robot mainly relies on the force control during contact to regulate its global position and orientation), \textit{(2)} significant dynamics model error and uncertainty that are inherently hard to model accurately, especially with contact and hardware load during extreme motions, and \textit{(3)} the trade-off between model accuracy and efficient computation for real-time execution. Achieving both target accuracy and robustness for long-flight maneuvers, such as high or long jumps, therefore still presents an open challenge. 
In this work, we address this challenge by developing a real-time MPC for quadruped jumping using a robot dynamics model learned from experiments.





\rebut{Many control and optimization techniques have been developed for jumping motions.}\redst{Various dynamics models have been proposed for control and optimization of aggressive robot maneuvers.} \rebut{Trajectory optimization (TO) with full-body nonlinear dynamics is normally utilized to generate long-flight trajectories in an offline fashion (e.g., \cite{QuannICRA19,chuongjump3D}).}
\redst{Many frameworks sacrifice model accuracy for very efficient computation by using simplified models such as spring-loaded inverted pendulum \cite{Patrick14_jump_humanoid} or single-rigid-body dynamic (SRBD) \cite{NMPC_3D_hopping, ZhitaoIROS22, GabrielICRA2021, park2017high,YanranDingTRO} that lumps the trunk and legs into a unified body.}
\rebut{Many MPC approaches sacrifice model accuracy to achieve robust maneuvers by using simplified models that treat the trunk and legs as a unified body \cite{NMPC_3D_hopping,ZhitaoIROS22,GabrielICRA2021, park2017high,YanranDingTRO}. 
Our recent iterative learning control (ILC) work \cite{chuong_ilc_jump} handles model uncertainty to realize long-flight jumps via multi-stage optimization that optimizes the control policy after each jump offline until reaching a target accurately. It also relies on a simplified model for computational efficiency. However, this work focuses on target jumping rather than robustness, e.g. requiring the same initial condition for all trials. Different from existing works, we learn a robot dynamics model from experiments and develop real-time MPC using the learned dynamics to achieve both target accuracy and robustness in continuous quadruped jumping.}
\redst{Recently, designing models that both consider leg dynamics and enable real-time execution has gained considerable attention. Zhou \textit{et al.} \cite{ZiyiZhouRAL} propose a linearized variant of the full centroidal dynamics, incorporating joint information in a centroidal model of quadruped jumping. He \textit{et al.} \cite{He2024} exploit a centroidal inertia decomposition model to regulate angular momentum during the flight phase of bipedal jumping. However, developing a model that can account for model mismatch with hardware and dynamics uncertainty as well as effectively predict long flight phases, which is important for control policies to enhance both target accuracy and jumping robustness, is still an open challenge.}
\begin{figure}[!t]
    \centering
    \includegraphics[width=\linewidth,trim={0cm 0cm 0cm 0cm},clip]{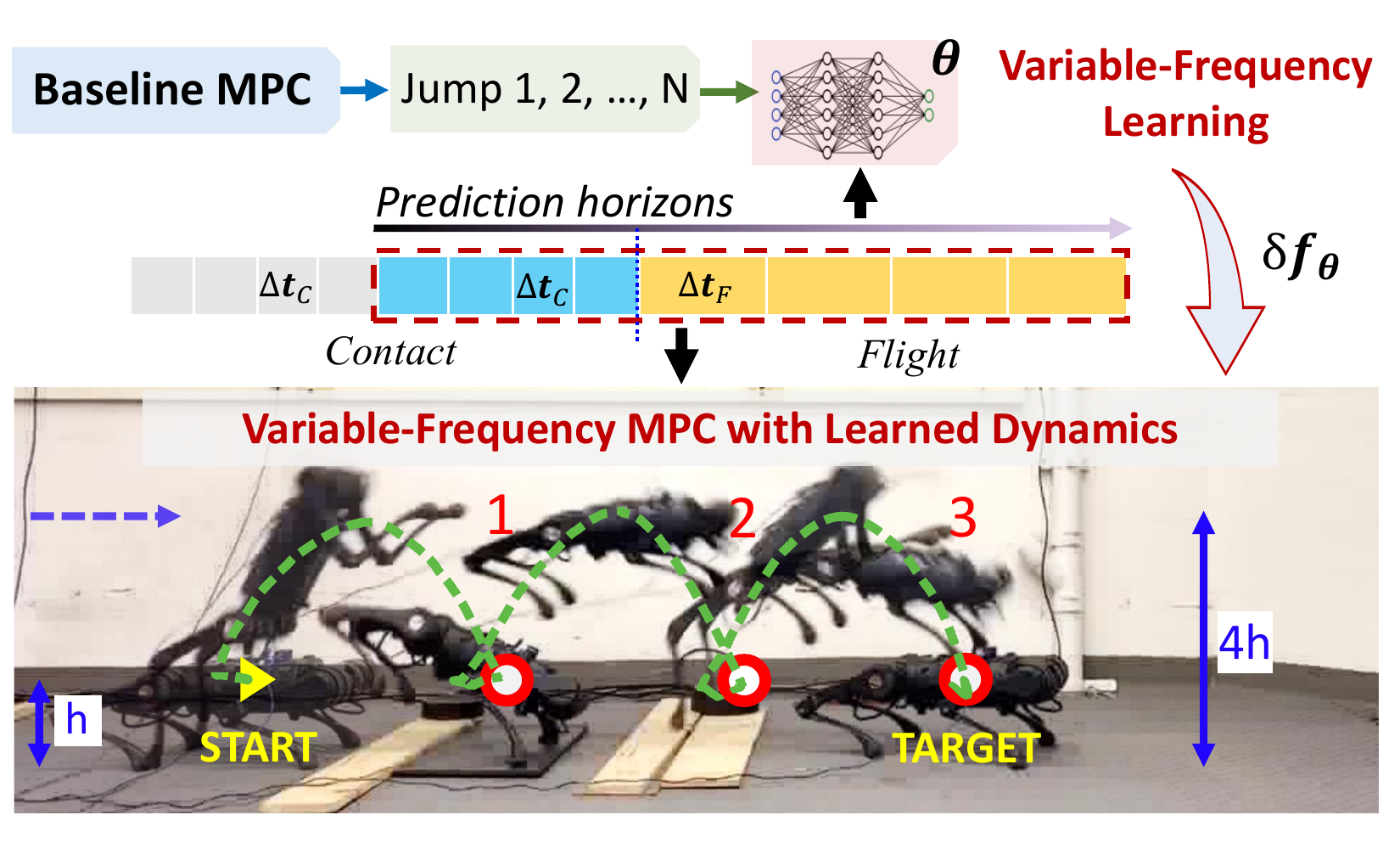}%
    \caption{A Unitree A1 robot performs \textit{continuous jumps} on unknown uneven terrain, achieving both target accuracy and robustness. The target distance for each jump is $0.6$ m. The flight phase covers a vertical height of up to $4 \times$ the robot's normal height. \rebut{MPC with a nominal single rigid-body model is used to collect data for training a neural-network residual dynamics model following using a variable-frequency scheme. The learned model is then used in a variable-frequency MPC to execute jumping motions. The variable-frequency scheme varies the time step sizes in the contact phase ($\Delta t_{\calC}$) and flight phase ($\Delta t_{\calF}$) to dedicate more model capacity and more MPC optimization steps to the contact phase, which improves the jumping robustness and accuracy. The green dashed line is the actual robot trajectory.} Supplemental video: \url{https://youtu.be/yUqI_MBOC6Q}.}
    \label{fig:introduction}
\vspace{-0.6cm} 
\end{figure}
Learning robot models from experiments to capture complex dynamics effects and account for model errors with real hardware has become a popular approach \cite{ bauersfeld2021neurobem,hewing2019cautious,saviolo2022physics,duong23porthamiltonian}.
Many frameworks learn residual dynamics using neural networks \cite{bauersfeld2021neurobem, salzmann2023real} or Gaussian processes \cite{hewing2019cautious, cao2017gaussian} and use MPC to control the learned systems \cite{salzmann2023real, pohlodek2022hilompc} to improve tracking errors.
Most existing works, however, primarily investigate dynamical systems without contact. Recently, Sun et al. \cite{sun2021online} proposed a notable method to learn a linear residual model, addressing unknown dynamics during walking stabilization.
\redst{Nevertheless, the method is designed for walking stabilization specifically and does not leverage the effect of leg dynamics.} 
\rebut{Pandala et al. \cite{Pandala_robustMPC}  propose to close the gap between reduced- and full-order models by using deep reinforcement learning to learn unmodeled dynamics, which are used in MPC to realize robust walking on various terrains. Meanwhile, we use supervised learning to learn unmodelled dynamics with real data from hardware.} 
\redst{Compared to existing works, learning dynamics for aggressive maneuvers with long-flight periods on legged robots presents some challenges,}
\rebut{Also, we learn dynamics for maneuvers with long-flight periods to tackle} \textit{(1)} the switching between multiple dynamic models due to contact, including flight phase where control action has very little effect on the body dynamics, \textit{(2)} disturbances and uncertainty in dynamic modeling due to hard impact in jumping, and \textit{(3)} the effect of intermittent control at contact transitions on state predictions.

In this letter, we propose a residual learning model that uses a \rebut{variable-frequency} \redst{adaptive-frequency} scheme, \rebut{i.e., varying the coarseness of integration time step}, to address the aforementioned challenges and enhance long-term trajectory predictions over different jumping phases. MPC has been commonly used for jumping by optimizing the control inputs during the contact phase based on future state prediction during the flight phase (e.g.,\cite{GabrielICRA2021, fullbody_MPC, park2017high, YanranDingTRO,  continuous_jump}). 
A major challenge in MPC for long flight maneuvers is to utilize a limited number of prediction steps yet still effectively cover the entire flight phase and especially the final jumping target. Another challenge is to obtain an accurate model for complex dynamic maneuvers, unknown dynamics, and model mismatch with the real hardware to improve the jumping accuracy, while still ensuring real-time performance. Some recent works have addressed these challenges partially.
For example, many MPC methods utilize conventional single-rigid-body dynamic (SRBD) models, ignoring the leg dynamics, to achieve real-time execution \cite{GabrielICRA2021,continuous_jump_bipedal,YanranDingTRO}. 
\rebut{Using the SRBD model in the contact phase can lead to inaccurate predictions of the robot's state at take-off, which is the initial condition of the
projectile motion in the flight phase. Thus, the trajectory prediction error can accumulate significantly over long flight periods.}
Some methods account for leg inertia \cite{ZiyiZhouRAL,He2024,fullbody_MPC}, however, disturbance and uncertainty in dynamic modeling have not been considered. 
\rebut{Recently, planning with multi-fidelity models and multi-resolution time steps has emerged as an effective strategy to enhance target accuracy and robustness \cite{heli_cafempc,Norby_adaptivempc,Heli_hierarchympc}. Li et al. \cite{Heli_hierarchympc} adopt a less accurate model in the far horizon and a more accurate but expensive model in the near future. Norby et al. \cite{Norby_adaptivempc} adapt model complexity based on the task complexity along the prediction horizon. Li et al. \cite{Heli_hierarchympc} also design multi-resolution time steps to cover the whole flight phase. While we vary the coarseness of the time step as \cite{Heli_hierarchympc}, we design a novel residual model learning approach combined with variable-frequency MPC to address the aforementioned challenges.}
\redst{In this work, we address these challenges by learning a dynamics model from date and designing an adaptive-frequency MPC capturing both the contact and entire flight phase in the optimization process}

\textbf{Contributions:} The contributions of this letter are summarized as follows. 
\begin{itemize}
  \item We learn a \textit{residual dynamics model} directly from a small real experiment dataset. The model accounts for nonlinear leg dynamics, modeling errors, and unknown dynamics mismatch in hardware and during contact.
  
  

    \item We propose learning the model in an \rebut{\textit{variable-frequency scheme}}\redst{adaptive-frequency scheme} that leverages different time resolutions to capture the entire flight phase, the jumping target, and the contact transitions over a few-step horizon, thereby significantly improving the accuracy of long-term trajectory prediction.
    

  \item We develop  \rebut{\textit{variable-frequency MPC using the learned model}}\redst{adaptive-frequency MPC using the learned model} to synthesize controls that improve both target accuracy and robustness in dynamic robot maneuvers.
  
  \item \textit{Extensive hardware experiments} validate the effectiveness and robustness of our approach with single and consecutive jumps on uneven terrains. Comparisons with other MPC techniques using a nominal model or fixed-frequency are also provided.
\end{itemize}

\redst{Our approach outperforms other MPC settings, reducing the jumping distance error up to $8$ times. We demonstrate robust and consecutive jumping on uneven terrain with height perturbations ($13-27\%$ of the robot's standing height), achieving a distance error of $1.67$ cm on average. Testing on $34$ single and continuous jumps, our approach achieves target jumping error of less than $2$ cm ($3.5\%$) for different targets and uneven terrain.}

\begin{figure*}
      \centering
      \includegraphics[trim = 0mm 11mm 0mm 0mm, clip, width=\linewidth]{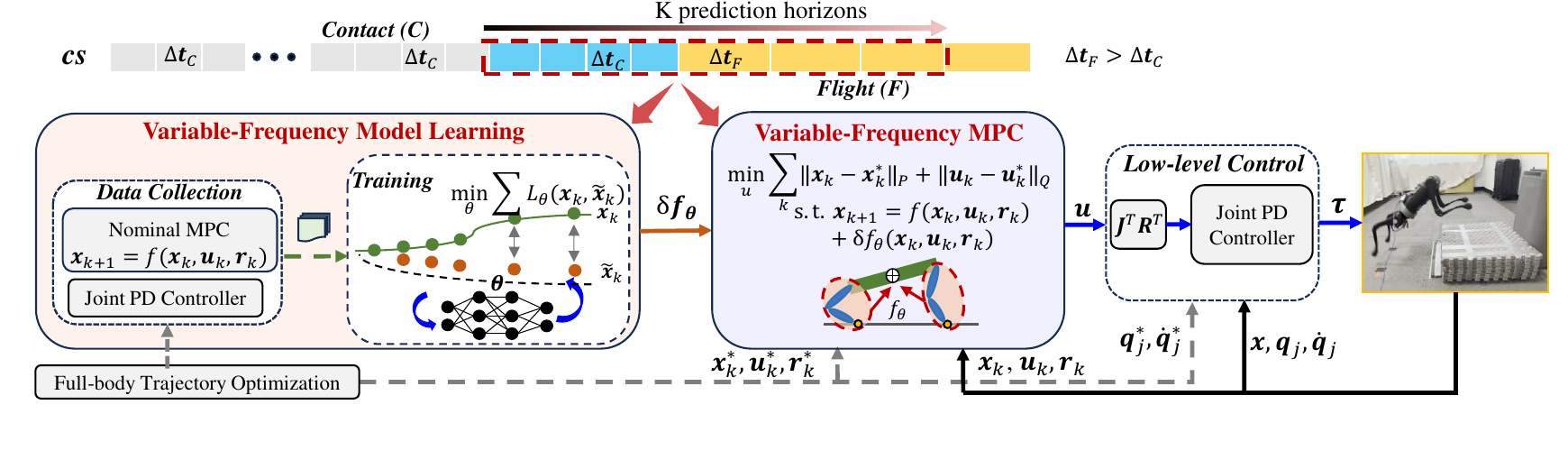}
      \caption{\textbf{System Architecture}. The learning procedure and MPC execution are paired with the same \rebut{integration timesteps} $\lbrace \Delta t_{\calC}, \Delta t_{\calF} \rbrace$, prediction horizon $K$, and predefined contact schedule ($cs$). The MPC and low-level joint PD controllers are updated at $40Hz$ and $1kHz$, respectively. 
      }
      \label{fig:overview_framework}
      \vspace{-1.7em}
\end{figure*}

\section{Problem Statement} \label{sec:problem_statement}


Consider a legged robot modelled by state $\bfx_k$, consisting of the pose and velocity of the body's center of mass (CoM), foot positions $\bfr_k$ relative to the body's CoM, and ground force control input $\bfu_k$ at the legs, sampled at time $t_k$. We  augment the nominal SRBD model (e.g., \cite{GabrielICRA2021, park2017high,YanranDingTRO}) with a learned residual term to account for leg dynamics, model mismatch with the real robot hardware, as well as capture complex dynamic effects in contact and hardware:
\rebut{
\begin{equation}\label{eq:discrete_learned_model}
    \bfx_{k+1} = \bff\left(\bfx_k, \bfu_k, \bfr_{k}, \Delta t_k \right) + \delta\bff_{\bftheta}\left(\bfx_k, \bfu_k, \bfr_{k}\right)\Delta t_k ,
\end{equation}
}
where $\bff$ represents the nominal SRBD model, $\delta\bff_{\bftheta}$ with parameters $\bftheta$ approximates the residual dynamics, and $\Delta t_k$ is the sampling interval. Our first objective is to learn the residual dynamics $\bff_{\bftheta}$ using a dataset of jumping trajectories.


\begin{problem} \label{problem:dynamics_learning}
Given a set $\calD \!=\! \{t_{0:K}^{(i)}, \bfx^{(i)}_{0:K}, \bfr^{(i)}_{0:K}, \bfu^{(i)}_{0:K}\}_{i=1}^D$ of $D$ sequences of states, foot positions, and control inputs,
find the parameters $\bftheta$ of the residual dynamics $\delta \bff_{\bftheta}$ in \eqref{eq:discrete_learned_model} by rolling out the dynamics model and minimizing the discrepancy between the predicted state sequence $\tilde{\bfx}^{(i)}_{1:K}$ and the true state sequence $\bfx^{(i)}_{1:K}$ in $\calD$:
\begin{align}
\min_{\bftheta} \;&\sum_{i = 1}^D \sum_{k = 1}^K \mathcal{L} (\bfx^{(i)}_k,\tilde{\bfx}^{(i)}_k) + \calL_{\text{reg}}(\bftheta) \notag\\
\text{s.t.} \;\; & \tilde{\bfx}^{(i)}_{k+1} = \bff \left(\tilde{\bfx}^{(i)}_k, \bfu^{(i)}_k, \bfr^{(i)}_k, \Delta t_k\right)\notag\\ 
&\qquad \qquad\qquad + \rebut{\delta \bff_{\bftheta}\left(\tilde{\bfx}^{(i)}_k, \bfu^{(i)}_k, \bfr^{(i)}_k,\right) \Delta t_k } , \notag\\
& \tilde{\bfx}^{(i)}_0 = \bfx_{0}^{(i)}, \quad \rebut{\forall i=1,...,D,}  \label{eq:problem_formulation_unknown_env_equation}
\end{align}
where $\mathcal{L}$ is an error function in the state space, and $\calL_{\text{reg}}(\bftheta)$ is a regularization term to avoid overfitting. Note that it is not necessary for $\Delta t_k = t_{k+1}-t_{k}$ to be fixed.
\end{problem}

After learning to improve the accuracy of model prediction, our second objective is to use the learned model \eqref{eq:discrete_learned_model} for MPC to track a desired state trajectory $\bfx_{0:K}^*$.

\begin{problem} \label{problem:mpc_learn}
Given the dynamics model \eqref{eq:discrete_learned_model}, \rebut{a current robot state $\bfx_0$} and foot positions $\bfr_0$, a desired trajectory of states $\bfx^*_{0:K}$, foot positions $\bfr^*_{0:K}$, and control $\bfu^*_{0:K}$,
design a control law $\bfu_0 = \bfpi(\bfx_0, \bfr_0, \bfx^*_{0:K}, \bfr^*_{0:K},\bfu^*_{0:K}; \bftheta)$ to achieve accurate and robust tracking of the desired state trajectory via shifting-horizon optimization:
\begin{align}
\min_{\bfu_{0:K-1}} \;& \sum_{k = 1}^K \| \bfx_k-\bfx_{k}^* \|_{\bfQ_k} + \| \bfu_{k-1}-\bfu_{k-1}^* \|_{\bfR_k} \notag\\
\text{s.t.} \;\; & \rebut{\bfx_{k+1} = \bff\left(\bfx_k, \bfu_k, \bfr_{k}, \Delta t_k \right) + \delta \bff_{\bftheta}\left(\bfx_k, \bfu_k, \bfr_{k}\right) \Delta t_k }, \notag\\
& \bfu_k \in \calU_k, \quad \forall k=0,...,K-1, \label{eq:mpc_formulation}
\end{align}
where $\bfQ_k$ and $\bfR_k$ are positive definite weight matrices,
and $\calU_{0:K-1}$ represents an input constraint set. 
\end{problem}

Achieving accurate target jumping with MPC requires not only an accurate model but also the coverage of the final state upon landing, which represents the accuracy of the jumping target. Using a fixed-frequency has shown to be efficient for locomotion tasks with limited aerial phases; however, this can face two challenges in long-flight maneuvers. On the one hand, using a fine time-step dirscretization enhances the model prediction accuracy but requires a large number of steps to capture the entire flight phase, thereby increasing the optimization size and computational cost. On the other hand, using a coarse time-step discretization allows capturing the entire flight phase efficiently but can sacrifice model prediction accuracy. For jumping tasks, different phases may require different model resolutions, e.g., fine time resolution during the contact phase but coarser time resolution during the flight phase as the model complexity is reduced due to unavailable force control and contact. Therefore, we propose to learn a model for dynamic maneuvers (Problem \ref{problem:dynamics_learning}) with an \rebut{variable-frequency scheme}\redst{adaptive-frequency scheme} that uses a coarse time discretization in the flight phase and a fine time discretization in the contact phase. Importantly, this \rebut{variable-frequency}\redst{adaptive-frequency} scheme is also synchronized with the MPC control of the learned model (Problem \ref{problem:mpc_learn}) \rebut{by utilizing the same time steps for both fitting the learned model and performing the MPC optimization. This synchronization ensures that the same discretization errors are leveraged when using the model for MPC predictions}. Thus, in our formulation $t_{0:K}$ does not have equal time steps and is a mixture of fine and coarse discretization capturing the contact and flight phases\redst{of a jumping motion}.
\rebut{Note that we use a constant time step $\Delta t_k$ during the different jumping phases. It is possible to use $\Delta t_k$ as input to the residual model neural networks. However, we need to collect a substantial amount of data for a wide range of $\Delta t_k$ to avoid overfitting, which would also increase the size of the neural network and computational cost.
}

\section{System Overview}
\label{sec:sys_overview}

Fig. \ref{fig:overview_framework} presents an overview of our system architecture. Our approach consists of two stages: \rebut{variable-frequency}\redst{adaptive-frequency} model learning (Sec. \ref{sec:adaptive-frequency-learning}) and \rebut{variable-frequency}\redst{adaptive-frequency} MPC (Sec. \ref{sec:adaptive-MPC}), which solve Problem~\ref{problem:dynamics_learning} and Problem~\ref{problem:mpc_learn}, respectively.

We synchronize the \rebut{variable-frequency scheme}\redst{adaptive-frequency scheme}  for both model learning and MPC execution with the same \textit{(1)} \rebut{variable prediction timesteps}\redst{adaptive sampling time} $\Delta t_{\calC}$ and $\Delta t_{\calF}$ for contact and flight phase, respectively, \textit{(2)} the same horizon length $K$ for all data collection, training, and MPC, and \textit{(3)} the same contact schedule.
Full-body trajectory optimization (TO) \cite{QuannICRA19} is utilized to generate jumping references for various targets, including body states $\bfx^*$, joint states $\lbrace \bfq_{\mathbf{J}}^*, \dot{\bfq}_{\mathbf{J}}^* \rbrace$, ground contact force $\bfu^*$, and foot positions $\bfr^*$. 
For data collection, we combine a baseline MPC using a nominal SRBD model and joint PD 
controller, generating diverse motions under disturbances. For training, we design a neural network to learn the discretized residual model $\delta \bff_{\bftheta}$ with \rebut{variable}\redst{adaptive} sampling time via supervised learning. For control, we design an \rebut{variable-frequency}\redst{adaptive-frequency} MPC using the learned dynamics to track a desired reference trajectory obtained from full-body TO. 
The feedback states from the robot include global body's CoM $\bfx$, and joint states $\bfq_{\mathbf{J}}, \dot{\bfq}_{\mathbf{J}} \in \mathbb{R}^4 $, and foot positions $\bfr \in \mathbb{R}^4$.

\section{\rebut{Variable-Frequency} Model Learning}
\label{sec:adaptive-frequency-learning}

In this section, we describe how to learn the residual dynamics $\delta \bff_{\bftheta}$ from data with \rebut{a variable-frequency}\redst{an adaptive-frequency} scheme that can cover the entire flight phase, the final state upon landing, and the contact transitions between jumping phases.



\subsection{Learning Dynamics with \rebut{Variable-Frequency}\redst{Adaptive Frequency}}  \label{sec:adaptive_learn_model}

We consider a 2D jumping motion on a legged robot with point foot contact, e.g., quadruped robot, with generalized coordinates $\boldsymbol{\mathfrak{q}} = \begin{bmatrix} \bfp^{\top} & \phi \end{bmatrix}^{\top} \!\in \mathbb{R}^{3}$, where $\bfp \in \bbR^2$ is the CoM's position, and $\phi \in \mathbb{R}$ is the body pitch angle. We define the generalized robot's velocity as $\bfzeta = \begin{bmatrix} \bfv^{\top} & \omega \end{bmatrix}^{\top} \in \mathbb{R}^3$, where $\bfv \in \mathbb{R}^2$ and $\omega \in \mathbb{R}$ are the linear and angular velocity. Both $\mathfrak{q}$ and $\bfzeta$ are expressed in world-frame coordinates. The robot state is $\bfx = \begin{bmatrix} \boldsymbol{\mathfrak{q}}^{\top} & \bfzeta^{\top} & g\end{bmatrix}^{\top} \in \mathbb{R}^{7}$, where the (constant) gravity acceleration $g$ is added to obtain a convenient state-space form \cite{Carlo2018}.
\rebut{We define $\bfR(\phi) = \begin{bmatrix}
    \text{cos}(\phi) & -\text{sin}(\phi); \text{sin}(\phi) & \text{cos}(\phi)
\end{bmatrix} \in \mathbb{R}^{2\times 2}$ as a rotation matrix of the main body, which converts $\bfr_{i,b}$ (a relative foot $i \in \{1, 2\}$ position relative to the body's CoM in the body frame) to the world frame $\bfr_{i} = \bfR\bfr_{i,b}$}. We denote $\bfr = \begin{bmatrix} \bfr_1^{\top} & \bfr_2^{\top} \end{bmatrix} ^{\top}\in \mathbb{R}^{4}$\redst{, where $\bfr_i$ as the foot $i \in \{1, 2\}$ position relative to the body's CoM in the world frame}. With the force control input for the front and rear legs as $\bfu = \begin{bmatrix} \bfu_f^{\top} & \bfu_r^{\top} \end{bmatrix} ^{\top}\in \mathbb{R}^{4}$, the nominal discrete-time SRBD model can be written as:
\begin{equation}\label{eq:discrete_nominal_model}
    \bff\left(\bfx_k, \bfu_k, \bfr_{k},\Delta t_k \right)  = \bfA_k\bfx_{k} + \bfB_k(\bfr_k) \bfu_k,
\end{equation}
where $\bfA_k = \bfI_{7} + \bfA_{ct} \Delta t_k$, $\bfB_k(\bfr_k)= \bfB_{ct}(\bfr_k) \Delta t_k$, $\Delta t_k$ is the \redst{adaptive sampling time}\rebut{time step} (e.g., $\Delta t_{\calC}$ or $\Delta t_{\calF}$ for the contact or flight phase, respectively), and $\bfA_{ct}$ and $\bfB_{ct}$ are obtained from the continuous-time robot dynamics:
%
\begin{equation}
    \bfA_{ct} = 
    \begin{bmatrix}
        \bf0 & \bfI_{3} & \bf0 \\ 
        \bf0 & \bf0 & \bfe_{3} \\ 
        \bf0 & \bf0 & 0
    \end{bmatrix}, 
    \bfB_{ct}(\bfr) = 
    \begin{bmatrix}
        \bf0 & \bf0 \\ 
        \bfI_{2} / m & \bfI_{2} / m \\ 
        [\bfr_1]_{\times}^{\top}/J & [\bfr_2]_{\times}^{\top}/J \\ 
        \bf0 & \bf0
    \end{bmatrix}, \nonumber
\end{equation}
where $\bfA_{ct}\in \mathbb{R}^{7\times 7}$, $\bfB_{ct} \in \mathbb{R}^{7\times 4}$, $\bfe_3= \begin{bmatrix} 0 & -1 & 0
\end{bmatrix}^\top$, $m$ and $J$ are mass and moment inertial of the body\redst{$m$ is the body mass, $J$ is the moment inertia of the body}, $[\bfr_i]_{\times} = \begin{bmatrix} r_{iz} & -r_{ix} \end{bmatrix}^{\top} \in \mathbb{R}^2$.
The residual term $\delta\bff_{\bftheta}\left(. \right)$ in (\ref{eq:discrete_learned_model}) is\redst{defined as}
\begin{equation}\label{eq:discrete_learned_res_model}
\delta\bff_{\bftheta}\left(. \right) = 
 \bfh_{\bftheta}(\bfx_k, \bfr_{k}) \Delta t_k  + \bfG_{\bftheta}(\bfx_k, \bfr_{k}) \Delta t_k \bfu_k,
\end{equation}
where $\bfh_{\bftheta}(\bfx_k, \bfr_{k})$ and $\bfG_{\bftheta}(\bfx_k, \bfr_{k})$ are represented by neural networks with learning parameters $\bftheta$.
Since $\bfu_k=\bf0$ during the flight phase, $\bfG_{\bftheta}\bfu_k=\bf0$. We have two separate models for the contact ($\calC$) and flight phase ($\calF$):
\begin{subequations}\label{eq:res_dynamics_each_phase}
    \begin{align}
     & (\calC):  ~ \bfx_{k+1} = \bfA_{\calC}\bfx_k + \bfB_{\calC}(\bfr_k) \bfu_k \nonumber \\
     & ~~~~~~~~~ + \bfh_{\bftheta_1}(\bfx_k, \bfr_{k})  \Delta t_{\calC} + \bfG_{\bftheta_1}(\bfx_k, \bfr_{k})\Delta t_{\calC} \bfu_k, \\ 
     &(\calF):  ~~~ \bfx_{k+1} = \bfA_{\calF}\bfx_k + \bfh_{\bftheta_2}(\bfx_k, \bfr_{k})\Delta t_{\calF}
     \end{align}
\end{subequations}
where $\bfA_{\calC} = \bfI_7 + \bfA_{ct} \Delta t_{\calC}$, $\bfB_{\calC}(\bfr_k)=\bfB_{ct}(\bfr_k) \Delta t_{\calC}$, and $\bfA_{\calF} = \bfI_7 + \bfA_{ct} \Delta t_{\calF}$.
We roll out the dynamics based on \eqref{eq:res_dynamics_each_phase}, starting from initial state $\bfx_0$ with given control input sequence $\bfu_{0:K-1}$ to obtain a predicted state sequence $\tilde{\bfx}_{1:K}$. Using the \rebut{variable-frequency}\redst{adaptive-frequency} scheme, the state prediction accounts for contact transitions (feet taking off the ground), a long flight phase, and the final robot state upon landing. We define the loss functions in Problem \ref{problem:dynamics_learning} as follows:
\begin{equation}
\begin{aligned}
\mathcal{L} (\bfx,\tilde{\bfx}) & = \| \boldsymbol{\mathfrak{q}} - \tilde{\boldsymbol{\mathfrak{q}}}\|^2_2 + \| \bfzeta - \tilde{\bfzeta}\|^2_2\\
\mathcal{L}_{reg} (\bftheta) & = \alpha_1 \Vert  \bfh_{\bftheta} \Vert + \alpha_2 \Vert  \bfG_{\bftheta}  \Vert.
\end{aligned}
\end{equation}
The parameters $\bftheta = [\bftheta_1, \bftheta_2]$ for each phase are updated by gradient descent to minimize the total loss.

\subsection{Data Collection}
\label{subsec:data_gen}

\rebut{For model learning, we directly collect state-control trajectories from hardware experiments by implementing}\redst{To collect state-control trajectories for residual dynamics learning, we implemented} an MPC controller with the nominal dynamics model \eqref{eq:discrete_nominal_model} and a reference body trajectory $\bfx^*$ obtained from full-body TO.
The TO assumes jumping from flat and hard ground with point foot contact. We generated various jumps \redst{trajectories} to different targets under different \redst{ground}disturbances (e.g., blocks under the robot feet with random height) to obtain a diverse dataset.

While the MPC aims to track the body reference trajectory $\bfx_{0:K}^*$, a joint PD controller is used to track the joint reference trajectory $(\bfq_{\mathbf{J}}^*, \mathbf{\dot q}_{\mathbf{J}}^*)$ from the full-body TO via  $\bftau_{pd,setpoint} = \bfK_{p}(\bfq_{\mathbf{J}}^*-\bfq_{\mathbf{J}})+\bfK_{d}(\mathbf{\dot q}_{\mathbf{J}}^*-\mathbf{\dot q}_{\mathbf{J}})$. Thus, the evolution of the robot states is governed by the combination of MPC and the joint PD controller. We collected the trajectory dataset $\calD$ with inputs $\bfu =  \bfu_{mpc} + \bfu_{pd},$
where $\bfu_{pd} =\left(\mathbf{\mathbf{J}}(\bfq_{\mathbf{J}})^{\top} \bfR^{\top}\right)^{-1} \bftau_{pd,setpoint}$, \redst{with rotation matrix $\bfR(\phi) = \begin{bmatrix}
    \text{cos}(\phi) & -\text{sin}(\phi); \text{sin}(\phi) & \text{cos}(\phi)
\end{bmatrix} \in \mathbb{R}^{2\times 2}$ and }$\mathbf{J}(\bfq_{\mathbf{J}})$ is the foot Jacobian.\redst{at joint configuration $\bfq_{\mathbf{J}}$}. 

The dataset is collected at the different time steps for the contact and flight phases, i.e., $\Delta t_{\calC}$ and $\Delta t_{\calF}$, respectively. \rebut{The data for each jump is then chunked by shifting a sliding window, size of $K$ by 1 timesteps}. Let $N$ be the number of collected state-control data points for each jump and $H$ be the number of jumps. We then obtain $D = H\times(N-K+1)$ state-control trajectories in total.

\section{\rebut{Variable-Frequency}\redst{Adaptive-Frequency} MPC with Learned Dynamics} \label{sec:adaptive-MPC}

In this section, we design \rebut{a variable-frequency}\redst{an adaptive-frequency} MPC controller for the learned dynamics \eqref{eq:res_dynamics_each_phase} to track a desired jumping reference trajectory obtained from full-body TO. \rebut{For a given robot state $\bfx_0$,} we formulate MPC as:
%
\begin{subequations} \label{eq:MPC_learned}
\begin{align}
 \min_{\bfu_{0:K-1}} & \; \sum_{k=1}^{K} \| \bfx_k-\bfx_{k}^* \|_{\bfQ_k} + \| \bfu_{k-1}-\bfu_{k-1}^* \|_{\bfR_k},\\
\text{s.t. }
     & (\calC):  ~~ \bfx_{k+1} = \bfA_{\calC}\bfx_k + \bfB_{\calC}(\bfr_k) \bfu_k \nonumber  \\ 
     & ~~~~~+ \bfh_{\bftheta_1}(\bfx_k, \bfr_{k})  \Delta t_{\calC} + \bfG_{\bftheta_1}(\bfx_k, \bfr_{k})\Delta t_{\calC} \bfu_k, \\ 
     &(\calF):  ~~ \bfx_{k+1} = \bfA_{\calF}\bfx_k + \bfh_{\bftheta_2}(\bfx_k, \bfr_{k})\Delta t_{\calF}, \\
     & \underline{\bfc}_k \leq \bfC_k \bfu_k \leq \bar{\bfc}_k, \; \forall k=0, ..., K-1, \label{eq:learn_MPC_ineq} \\ 
     & \bfD_k \bfu_k =\bm{0}, \qquad\quad \forall k=0, ..., K-1, \label{eq:learn_MPC_eq} 
\end{align}
\end{subequations}
where \eqref{eq:learn_MPC_ineq} represents input constraints related to friction cone and force limits and \eqref{eq:learn_MPC_eq} aims to nullify the force on the swing legs based on the contact schedule.

With the MPC horizon including $K_{\calC}$ steps in the contact phase and $K_{\calF} = K-K_{\calC}$ steps in the flight phase, we define \rebut{$\tilde{\bfU}_{\calC} = [\tilde{\bfu}_0^{\top}, \tilde{\bfu}_{1}^{\top}, ..., \tilde{\bfu}_{K_{\calC} - 1}^{\top}]^{\top} \in \mathbb{R}^{K_{\calC} \times 4}$}
%
%
as a concatenation of the control inputs in the contact phase. With this notation, the predicted trajectory is:
%
\begin{subequations}\label{eq:adaptive_prediction}
    \begin{align}
        (\calC): ~ \tilde{\bfX}_{\calC}&= \calA_\calC \bfx_0 + (\calB_{\calC} + \calB_{\calC, \bftheta} ) \tilde{\bfU}_{\calC} + \bfH_{\calC, \bftheta}\label{eq:predict_phase1}\\
        (\calF): ~ \tilde{\bfX}_{\calF}&= \calA_{\calF} \tilde{\bfx}_{K_c}+ \bfH_{\calF, \bftheta} \label{eq:predict_phase2}
     \end{align}
\end{subequations}
%
where
$\tilde{\bfX}_{\calC} = [\tilde{\bfx}_{1}^{\top}, \tilde{\bfx}_{2}^{\top}, ..., \tilde{\bfx}_{K_c}^{\top}]^{\top} $ and 
$\tilde{\bfX}_{\calF} = [\tilde{\bfx}_{K_c+1}^{\top}, \tilde{\bfx}_{K_c+2}^{\top}, ..., \tilde{\bfx}_{K}^{\top}]^{\top} $ denote the concatenation of predicted states belonging to the contact and flight phase, respectively. The matrices $\calA_{\calC}$ and $\calA_{\calF}$ are computed as
%
\rebut{
\[\left\{
  \begin{array}{lr}
    \calA_{\calC}= \begin{bmatrix}
(\bfA_{\calC})^\top
(\bfA_{\calC}^2)^\top
\ldots 
(\bfA_{\calC}^{K_c})^\top
\end{bmatrix} \in \mathbb{R}^{7 K_c \times 7},  \\
\calA_{\calF}= \begin{bmatrix}
(\bfA_{\calF})^\top
(\bfA_{\calF}^2)^\top
\ldots 
(\bfA_{\calF}^{K_{\calF}})^\top
\end{bmatrix} \in \mathbb{R}^{7 K_{\calF} \times 7}.
\end{array}
\right.
\]
}
\rebut{While we use the current robot state for training the neural networks $\lbrace \bfG_{\bftheta}$, $\bfh_{\bftheta}\rbrace$ as presented in Sec. \ref{sec:adaptive-frequency-learning}, we utilize the reference trajectory ($\bfx_{k}^*, \bfr_{k}^* $) in future MPC horizons for guiding the MPC since the future robot states are not available at each MPC update (\cite{continuous_jump_bipedal,YanranDingTRO, GabrielICRA2021,ZhitaoIROS22, park2017high}). 
The reference will be used to compute the output of the neural networks $\bfG_{\bftheta}$ and $\bfh_{\bftheta}$ in future horizons. }
\rebut{In particular, we still use the current robot state and foot position $(\bfx_0, \bfr_{0})$ to evaluate the residual term $\bfB_{\bftheta,0} = \bfG_{\bftheta}(\bfx_0, \bfr_{0}) \Delta t_0 $ for each MPC update, then define 
$\bfB_{\bftheta,k} = \bfG_{\bftheta}(\bfx_{k}^*, \bfr_{k}^*) \Delta t_k$, $k = 1,..., K_{\calC}$. }\redst{We estimate the residual terms via where $\bfx_{k}^*, \bfr_{k}^* $ are the reference states and foot positions, }
\rebut{In (\ref{eq:predict_phase1}), we obtain $[\calB_{\calC}]_{\lbrace m,n \rbrace} =   \bfA_{\calC}^{m-n} \bfB_{n-1}$ and its residual $\left[\calB_{\calC, \bftheta}\right]_{\lbrace m,n \rbrace} = \bfA_{\calC}^{m-n} \bfB_{\bftheta,n-1} ~ \textit{if} ~ m \geq n, \textit{and} ~ \bm{0} ~ \textit{otherwise}.$}
\redst{\begin{equation}
     \begin{array}{lr}
    [\calB_{\calC}]_{\lbrace m,n \rbrace} ~~=   \bfA_{\calC}^{m-n} \bfB_{n-1}  ~~~ \textit{if} ~ m \geq n, \textit{and} ~ \bm{0} ~ \textit{otherwise}  \\
   \left[\calB_{\calC, \bftheta}\right]_{\lbrace m,n \rbrace} = \bfA_{\calC}^{m-n} \bfB_{\bftheta,n-1} ~ \textit{if} ~ m \geq n, \textit{and} ~ \bm{0} ~ \textit{otherwise}
     \end{array}  \nonumber
\end{equation}}

\rebut{For the sake of notation, we define $\chi_k =(\bfx_k, \bfr_k)$, $\chi_k^* =(\bfx_k^*, \bfr_k^*)$ as the actual and reference robot state and foot trajectory. The residual matrix $\bfH_{\calC, \bftheta}$ in (\ref{eq:adaptive_prediction}) is $\bfH_{\calC,\bftheta} =  \Delta t_{\calC}$ $\mathbf{S}_{\calC} \left[
\bfh_{\bftheta}(\chi_0)^{\top}~
\bfh_{\bftheta}(\chi_1^*)^{\top}~...~
\bfh_{\bftheta}(\chi_{K_c-1}^*)^{\top}
\right]^{\top}$,} 
\redst{The residual matrix $\bfH_{\calC, \bftheta}$ in (\ref{eq:adaptive_prediction}) is $\bfH_{\calC,\bftheta} =  \Delta t_{\calC}$ $\mathbf{S}_{\calC} \left[
\bfh_{\bftheta}(\bfx_0, \bfr_{0})^{\top}~
\bfh_{\bftheta}(\bfx_{1}^*, \bfr_{1}^*)^{\top}~...~
\bfh_{\bftheta}(\bfx_{K_c-1}^*, \bfr_{K_c-1}^*)^{\top}
\right]^{\top}$,} 
\rebut{where $\bfS_{\calC}$ is lower triangular matrix with each entry $[\bfS_{\calC}]_{\lbrace m,n \rbrace} = \bfA_{\calC}^{m-n} \in \mathbb{R}^{7 \times 7}$, $m \geq n$. }\rebut{The residual matrix 
$\bfH_{\calF,\bftheta} = \Delta t_{\calF} \bfS_{\calF} \left[
\bfh_{\bftheta}(\chi_{K_c}^{*})^{\top} \quad
\ldots \quad 
\bfh_{\bftheta}(\chi_{K}^*)^{\top}
\right]^{\top}$}
\redst{$\bfH_{\calF,\bftheta} = \Delta t_{\calF} \bfS_{\calF} \left[
\bfh_{\bftheta}(\bfx_{K_c}^{*}, \bfr_{K_c}^*)^{\top}~...~ 
\bfh_{\bftheta}(\bfx_{K}^*, \bfr_{K}^*)^{\top}
\right]^{\top}$}
\rebut{where $\bfS_{\calF}$ is lower triangular matrix with each entry defined $[\bfS_{\calF}]_{\lbrace m,n \rbrace} = \bfA_{\calF}^{m-n} \in \mathbb{R}^{7 \times 7}$, $m \geq n$.} 
The state prediction at the end of $K_{\calC}$ steps can be computed as
\begin{equation}\label{eq:x_kNc}
    \tilde{\bfx}_{K_c}= \bfA_{\calC}^{K_c} \bfx_0 + (\bfS + \bfS_{\theta}) \tilde{\bfU}_{\calC} + \bm{\phi}_{\bftheta}, 
\end{equation}
where $\bfS = \left[\bfA_{\calC}^{K_c-1}\bfB_0 \quad \bfA_{\calC}^{K_c-2}\bfB_{1} \quad ... \quad \bfB_{K_c-1}  \right] $, $\bfS_{\bftheta} = \left[\bfA_{\calC}^{K_c-1}\bfB_{\bftheta,0} \quad \bfA_{\calC}^{K_c-2}\bfB_{\bftheta,1} \quad ... \quad \bfB_{\bftheta,K_c-1}  \right] $,  and $\bm{\phi}_{\bftheta} = \left[\bfA_{\calC}^{K_c-1}\bfh_{\bftheta,0} \quad \bfA_{\calC}^{K_c-2}\bfh_{\bftheta,1} \quad ...\quad \bfh_{\bftheta,K_c-1}  \right]$.\redst{Substituting \eqref{eq:x_kNc} into \eqref{eq:predict_phase2} yields:} \rebut{Substituting \eqref{eq:x_kNc} into \eqref{eq:adaptive_prediction} yields the state prediction as}
\begin{equation} \label{eq:state_prediction2}
    \tilde{\bfX}_{\calF}= \bfA_{\calF} \bfA_{\calC}^{K_c} \bfx_0+ \bfA_{\calF} (\bfS + \bfS_{\theta}) \tilde{\bfU}_{\calC}  + \bfA_{\calF} \bm{\phi}_{\bftheta} +  \bfH_{\calF, \bftheta}.
\end{equation}
\begin{equation}
    \tilde{\bfX} = \begin{bmatrix}
        \tilde{\bfX}_{\calC} \\ \tilde{\bfX}_{\calF}
    \end{bmatrix} = \bfA_{\text{qp}} + \left(\bfB_{\text{qp}} + \bfB_{\bftheta,\text{qp}}\right) \tilde{\bfU} + \bfH_{\bftheta}
\end{equation}

where, $\bfA_{\text{qp}} = \begin{bmatrix}
    \calA_{\calC} \\ \calA_{\calF} \bfA_{\calC}^{K_c}
\end{bmatrix}$, $\bfB_{\text{qp}} = \begin{bmatrix}
    \calB_{\calC} & \bf0 \\
    \calA_{\calF} \bfS & \bf0
\end{bmatrix}$, $\bfB_{\bftheta,\text{qp}} = \begin{bmatrix}
    \calB_{\calC,\bftheta} & \bf0 \\
    \calA_{\calF} \bfS_{\bftheta} & \bf0
\end{bmatrix}$, $\bfH_{\bftheta} = \begin{bmatrix}
    \bfH_{\calC, \bftheta} \\
    \bfH_{\calF, \bftheta} + \calA_{\calF} \bm{\phi}_{\bftheta}
\end{bmatrix}$, and $\tilde{\bfU} = \begin{bmatrix}
    \tilde{\bfU}_{\calC} \\
    \tilde{\bfU}_{\calF}
\end{bmatrix} $.

The objective of the MPC in \eqref{eq:MPC_learned} can be rewritten as:
\begin{equation}
    J(\bfU) = \| \tilde{\bfX}-\bfX^* \|_{\bfP}^2 + \| \tilde{\bfU}-\bfU^* \|_{\bfQ}^2,
\end{equation}
leading to a quadratic program (QP):
\begin{equation}\label{eq:MPC_learned_QPcompact}
\begin{aligned}
& \underset{\bfU}{\text{min}}
& & \bfU^{\top} \bfalpha \bfU + \bfbeta^\top \bfU  \quad \text{s.t.} \;\;\bfU_k \in \calU_k,
\end{aligned}
\end{equation}
where 
\begin{equation*}
    \begin{aligned}
        \bfalpha &= (\bfB_{\text{qp}}+ \bfB_{\bftheta,\text{qp}}) 
 ^{\top} \bfQ (\bfB_{\text{qp}}+ \bfB_{\bftheta,\text{qp}}) + \bfR, \\ 
        \bfbeta &= 2 (\bfB_{\text{qp}}+ \bfB_{\bftheta,\text{qp}}) 
 ^{\top} \bfQ \left( \bfA_{\text{qp}} \bfx_{k} + \bfH_{\bftheta}- \bfX^* \right) -\bfR \bfU^*,\\
        \calU_k &= \{ \bfu \in \bbR^4 \mid \underline{\bfc}_k \leq \bfC_k \bfu_k \leq \bar{\bfc}_k, \bfD_k \bfu_k =\bm{0}\}.
    \end{aligned}
\end{equation*}

\rebut{
}

\section{Evaluation}
\label{sec:results}

\subsection{Experiment Setup}

We evaluated our approach using a A1 robot. \rebut{To get data for real-time control, we utilized a motion capture system}\redst{we utilized an Optitrack motion capture system with five Optitrack Prime $13W$ cameras} for estimating the position and orientation of the robot trunk at $1$ kHz with position errors of $1$ mm. \rebut{We obtain velocity data from position data using forward finite-difference method.}
%

We used \textit{Pytorch} to implement and train the residual dynamics model. The trained \textit{Pytorch} model was converted to a torch script trace, which in turn, was loaded in our MPC implementation in C++. \rebut{The MPC is solved by a QP solver (qpOASES)}.
We utilized the optimization toolbox 
\rebut{\textit{CasADi}} \cite{casadi} to set up and solve the full-body TO for various jumping targets within $[0.3, 0.8]$ m on a flat ground. We considered jumping motions with contact phase (C) consisting of an all-leg contact period (e.g., $500$ ms) and a rear-leg contact period (e.g., $300$ ms). With the flight phase (F) scheduled for $400$ ms, the front legs and rear legs become swing legs (SW) up to $58\%$ and $33\%$ of the entire jumping period, respectively. Jumping maneuvers feature long flight phases, i.e., the robot can jump up to $4$ times its normal standing height during the mid-air period. \rebut{All jumping experiments are executed with sufficient battery level ($>90\%$)}

\subsection{Data Collection and Training}
\begin{table}[t]
	\centering
	\caption{Training Parameters}
	\begin{tabular}{ccc}
		\hline
		Parameter & Symbol & Value\\
		\hline
        \rebut{Variable time steps} & $\Delta t _{\calC}, \Delta t _{\calF} $ & $\lbrace 25, 100 \rbrace (ms)$ \\[.5ex]
        Prediction steps & $K$ & $10$ \\[.5ex]
		NN architecture & $\bfh_{\bftheta_1}(\mathbf\frakq, \bfr)$ & $7$ - $400$ Tanh - $400$ Tanh - $3$ \\[.5ex]
  		 & $\bfG_{\bftheta_1}(\mathbf\frakq, \bfr)$ & $7$ - $1000$ Tanh - $1000$ Tanh - $12$ \\[.5ex]
              & $\bfh_{\bftheta_2}(\mathbf\frakq, \bfr)$ & $7$ - $400$ Tanh - $400$ Tanh - $3$ \\[.5ex]
        Learning rate & $\gamma$ & $2.10^{-4}$ \\[.5ex]
        Regulation weights & $\alpha_1, \alpha_2$ & $10^{-3},10^{-3}$ \\[.5ex]
        Total training steps & $N_{train}$ & $2.10^{4}$ \\[.5ex]
		\hline 
		\label{tab:training_params}
	\end{tabular}
  \vspace{-2.2em}
\end{table}

\begin{figure}
\centering
\includegraphics[trim={0cm 0cm 0cm 0cm},clip,width=0.45\textwidth]{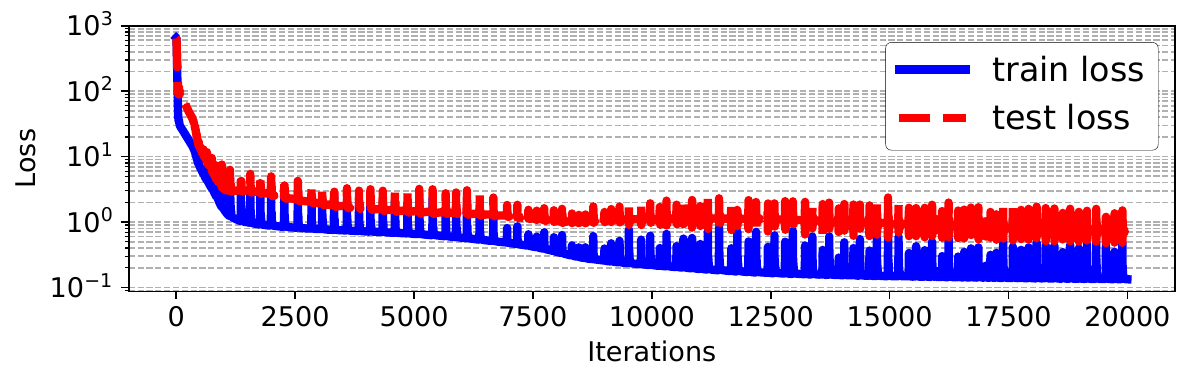}
\caption{\textbf{Training loss and testing loss (Log scale).}}
\label{fig:loss}
\vspace{-0.7cm} 
\end{figure}

A baseline MPC controller using the nominal model, described in Sec. \ref{subsec:data_gen}, was implemented to generate jumping trajectories under a variety of unknown disturbances (i.e. box under feet with random height within $[0, 8]$ cm) and \rebut{zero-height targets}.
We collected data from $H=20$ jumps with $80\%$ for training and $20\%$ for testing. The data points were sampled with variable sampling time  $\Delta t _{\calC} = 25$ ms, $\Delta t_{\calF} = 100$ ms, and horizon length of $K=10$. 
\redst{The dataset $\calD = \{t_{0:K}^{(i)}, \bfx^{(i)}_{0:K}, \bfr^{(i)}_{0:K}, \bfu^{(i)}_{0:K}\}_{i=1}^D$ consisted of $D = 20\times 28$ state-control sequences for training. }


The training parameters are listed in Table \ref{tab:training_params}. We used fully-connected neural networks with architectures listed in Table \ref{tab:training_params}: the first number is the input dimension, the last number is the output dimension, and the numbers in between are the hidden layers' dimensions and activation functions. 
\redst{For all networks, the input consists of body position, body orientation, and foot position w.r.t the body's CoM in the world frame. }\redst{Note that }The output of $\bfG_{\bftheta_1}$ is then converted to a $3\times4$ matrix. The training and testing losses are illustrated in Fig. \ref{fig:loss}. 




\begin{figure}
\centering
    \includegraphics[trim={0cm 0cm 0cm 0cm},clip,width=0.48\textwidth]{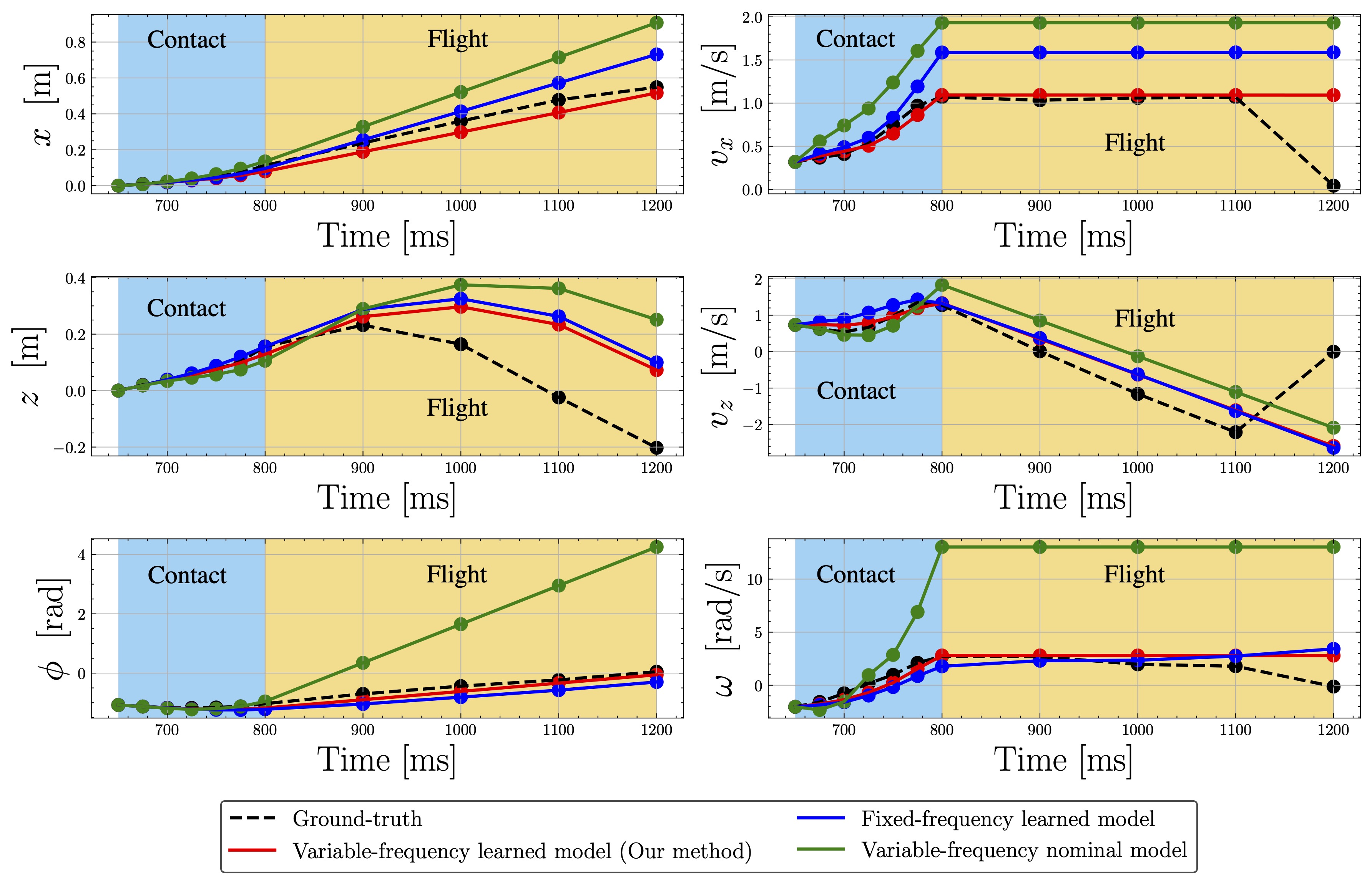}
    \caption{\textbf{State prediction comparison on a random testing dataset}.  \rebut{It starts at $650$ ms, consisting of $K_c = 6$ steps in the contact phase and $K_f = 4$ steps covering the flight period of $400$ ms. We compare variable-frequency learned model (our method), fixed-frequency learned model, variable-frequency nominal model, and the ground-truth trajectories. The ground-truth data is directly obtained from hardware experiments.}}
    \label{fig:roll_out_hybrid}
    \vspace{-0.3cm} 
\end{figure}
\begin{figure} 
    \centering
    {\centering
        \resizebox{1.1\linewidth}{!}{\includegraphics[trim={2.5cm 11.8cm 1cm 11.8cm},clip]{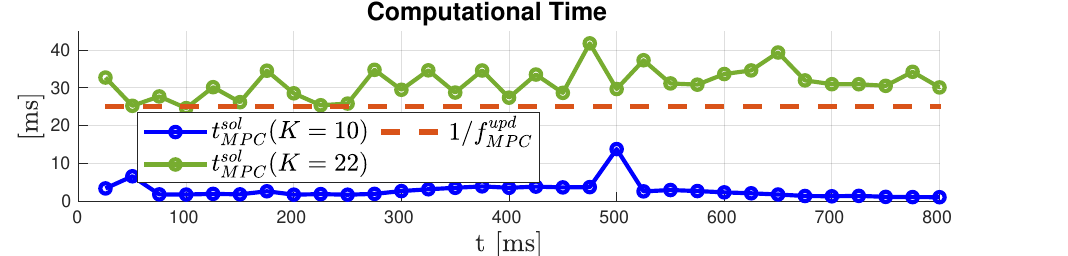}}}\\
    \caption{\rebut{MPC solving time ($t_{MPC}^{sol}$) with various horizons $K=10$ and $K=22$. The red dashed line denotes the MPC update time ($1/f_{MPC}^{upd}$) of $25$ ms. Real-time performance is achieved if $t_{MPC}^{sol}<<1/f_{MPC}^{upd}$.}}
    \label{fig:computational_time}
    \vspace{-0.2cm} 
\end{figure} 

\begin{figure}[!ht]
\centering
\includegraphics[trim={5cm 11cm 4cm 10.5cm},clip,width=0.5\textwidth]{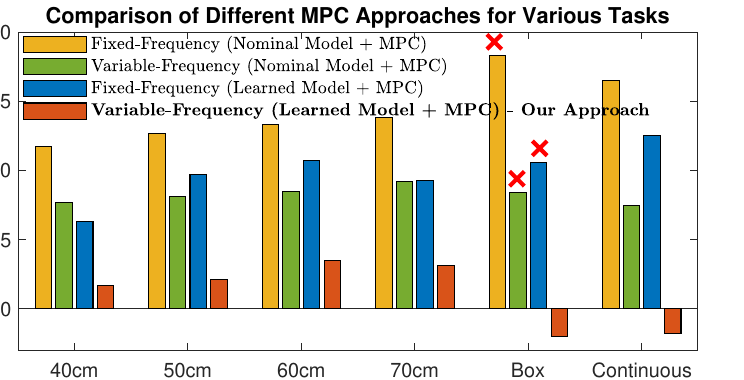}
    \caption{\rebut{Comparison of error of final target jumping (in cm) for four real-time MPC setups with different models and timesteps resolutions in a total of $92$ jumps with robot hardware. Positive values indicate that the jumps fall short of the target. The cross (\textcolor{red}{$\times$}) denotes failure to jump on the box. Supplemental video: \url{https://youtu.be/2QjZkARs1mU}.}}
	\label{fig:compare_approach}
     \vspace{-0.5cm} 
\end{figure}


\begin{figure*}[!t]
	\centering
	{\centering
		\resizebox{\linewidth}{!}{\includegraphics[trim={0cm 0.3cm 0cm 0cm},clip]{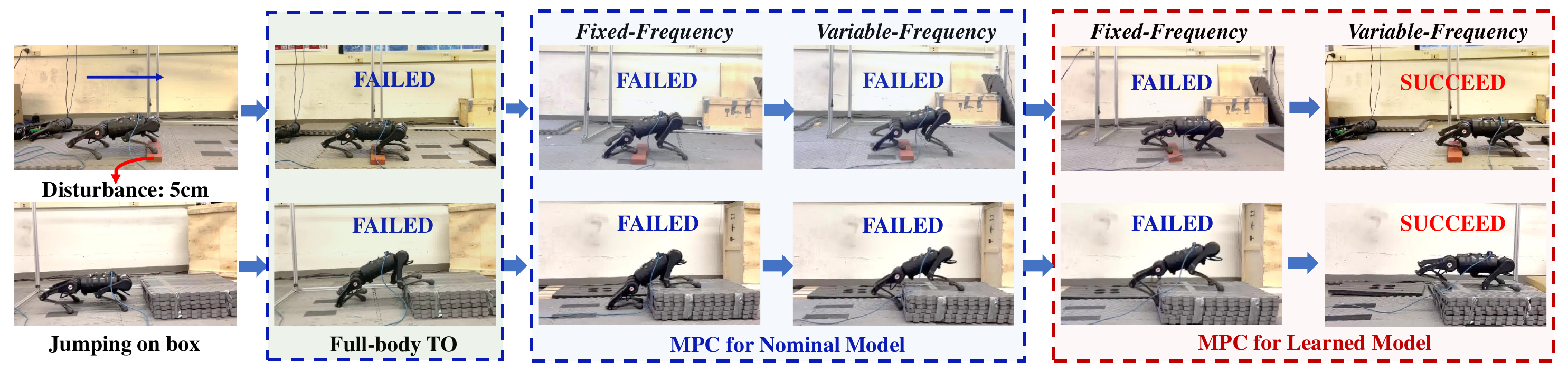}}
	}
	\caption{
	\textbf{Jumping from uneven terrain and Jumping on a box:} \rebut{Comparison among full-body TO, nominal-model MPC with fixed and variable frequency, and learned-model MPC with fixed and variable frequency. All MPCs have a horizon length of $K=10$. An unknown $5$ cm block ($30\%$ of the robot's initial height) was introduced under the robot's front feet (upper figures). With variable-frequency learned-model MPC (our method), the robot is able to compensate for the disturbance and perform successful jumps. Supplemental video:} \url{https://youtu.be/yUqI_MBOC6Q}.} 
\label{fig:box_jumping_and_forward60_dis}
    \vspace{-0.4cm} 
\end{figure*}

\subsection{Comparative Analysis}





\subsubsection{Evaluation on testing dataset}
We rolled out the learned dynamics with \rebut{variable frequency}\redst{adaptive frequency} to predict the future state trajectories and compared with \textit{(i)} those of \rebut{the variable-frequency nominal model}\redst{the nominal model with adaptive frequency}, \textit{(ii)} those of \rebut{the fixed-frequency learned model}\redst{the learned model with fixed frequency}, and \textit{(iii)} the ground-truth states in Fig. \ref{fig:roll_out_hybrid}. 
\redst{We select two \textit{random} subsets of testing data. The first starts at $525 (ms)$, leveraging a timing window of $K\Delta t_{\calC}$. This subset is within the scheduled contact phase as shown in Figure \ref{fig:roll_out_contact}. The second starts at $650 (ms)$, leveraging $K = 10$ horizons that consist of $K_c = 6$ horizon for contact phase and $K_f = 4$ horizon for flight phase, as depicted in Fig. \ref{fig:roll_out_hybrid}.} The fixed-frequency learned model is trained with $K=10$ and $\Delta t=25$ ms. The figure shows that our proposed learned model (red lines) outperforms variable-frequency nominal model (green lines) and fixed-frequency learned model (blue lines). 
\rebut{We highlight a large deviation in trajectory prediction of variable-frequency nominal model with the others. This inaccuracy can be explained by the use of conventional SRBD for the entire prediction horizon and the coarse Euler integration for the flight phase. Our variable-frequency learned model helps address the inaccuracies introduced by the nominal model and keeps sufficient accuracy of trajectory prediction while allowing real-time MPC by keeping a small number of decision variables.
}


\subsubsection{Effect of prediction horizon}
Fig. \ref{fig:computational_time} compares the solving time for MPC with different horizons. With our \rebut{variable-frequency scheme}\redst{adaptive-frequency scheme}, we can use a few prediction horizons, e.g., $K=10$ ($6$ for contact and $4$ for the flight phase), allowing QP solving time for each MPC update (\eqref{eq:MPC_learned_QPcompact}) to be $2$ ms only on average. This efficient computation enables real-time performance for MPC. Using a fixed-frequency scheme with sample time of $25$ ms requires MPC to use a large number of steps, e.g. $K=22$ ($6$ for contact and $16$ for the whole flight phase). \rebut{This long horizon yields a large optimization problem, which takes $\sim$$30$ ms for each MPC update and deteriorates the real-time performance.}
\redst{This long horizon yields a large optimization problem, causing significant computational costs. It takes an average of $30$ ms for each MPC update, thus deteriorating the real-time performance.}

\subsubsection{Execution}
We verified that our proposed approach enables both robust and accurate target jumping to various targets. We compared \textit{(a)} nominal model $\&$ fixed-frequency MPC, \textit{(b)} nominal model $\&$ \redst{adaptive-frequency}\rebut{variable-frequency} MPC, \textit{(c)} fixed-frequency learned model $\&$ fixed-frequency MPC, and \textit{(d)} \redst{adaptive-frequency}\rebut{variable-frequency} learned model $\&$ \redst{adaptive-frequency}\rebut{variable-frequency} MPC (\textit{our approach}). \rebut{The fixed-frequency scheme uses $(\Delta t _{\calC}, \Delta t _{\calF}) =(25, 25) ms$ and variable-frequency one utilizes $(\Delta t _{\calC}, \Delta t _{\calF}) =(25, 100) ms$. All MPCs use the prediction horizon $K=10$, thus the fixed-frequency MPC does not cover the entire flight phase. 

For each combination, we perform $23$ jumps consisting of (i) 3 single jumps for each of five targets: $x=40$ cm, $x = 50$ cm, $x = 60$ cm, $(x,z) = (60, 20)$ cm for jump on box, and $x = 70$ cm, and (ii) $8$ continuous jumps of $60$ cm. This yields a total of $92$ jumps for comparison. Fig.~\ref{fig:compare_approach} shows the average final jump target errors of four different MPC combinations across different jumping tasks.}
Our approach outperforms the methods that adopt fixed-frequency or nominal models, reducing the jumping distance error up to $8$ times.
\rebut{Our method demonstrates successful jumping on a box with \redst{an error of only $1-2$ cm in distance and} landing angle errors $<15^\circ$, while the fixed-frequency MPC or nominal model MPC fail, as shown in Fig.~\ref{fig:box_jumping_and_forward60_dis}. Note that the box-jumping task is not executed during data collection, verifying the scalability of our method to generalize to unseen reference trajectories and unseen tasks. With our method, the robot also successfully jumps under model uncertainty, e.g, an unknown $5cm$ block ($30\%$ robot standing height) placed under the front feet (Fig. \ref{fig:box_jumping_and_forward60_dis}). This task uses a reference trajectory designed for flat ground and demonstrates an example where the robot significantly deviates from the reference trajectory.}
\redst{Due to space limitation, we discuss two jumping tasks \textit{(i)} box jumping with the target at $(x,z)=(60,20)$ cm, and \textit{(ii)} forward jumping $60$ cm with an unknown disturbance (i.e., a block of $5$ cm under the robot's feet). The results are shown in Fig. \ref{fig:box_jumping_and_forward60_dis}} 
\redst{All MPC methods use prediction horizon $K=10$. The fixed-frequency scheme used a time step of $25$ ms for all jumping phases, so the MPC prediction does not cover the entire flight phase.}

We also studied the effect of uncertainty in the model by evaluating the jumping performance if we only utilize a joint PD controller to track the joint reference from full-body TO \cite{QuannICRA19}, via \rebut{ $\bftau_{pd} = \bftau_{ff} + \bfK_{p}(\bfq_{\mathbf{J}}^*-\bfq_{\mathbf{J}})+\bfK_{d}(\mathbf{\dot q}_{\mathbf{J}}^*-\mathbf{\dot q}_{\mathbf{J}})$}. The full-order model can be conservative by assuming hard feet, point-to-surface contact, and hard ground. These assumptions, however, are not valid for the Unitree A1 robot equipped with deformable feet \cite{Zachary_RAL_online_calibration_2022}.
Uncertainties due to DC motors working at extreme conditions or motor deficiency are difficult to model and are normally ignored in the full-body model. These factors affect the model accuracy and prevent the robot from reaching the jumping target if we solely rely on the joint PD, as can be seen in Fig. \ref{fig:box_jumping_and_forward60_dis}. 

\begin{figure}[t]
    \centering
    \includegraphics[width=\linewidth,trim={0cm 0cm 0cm 0cm},clip]{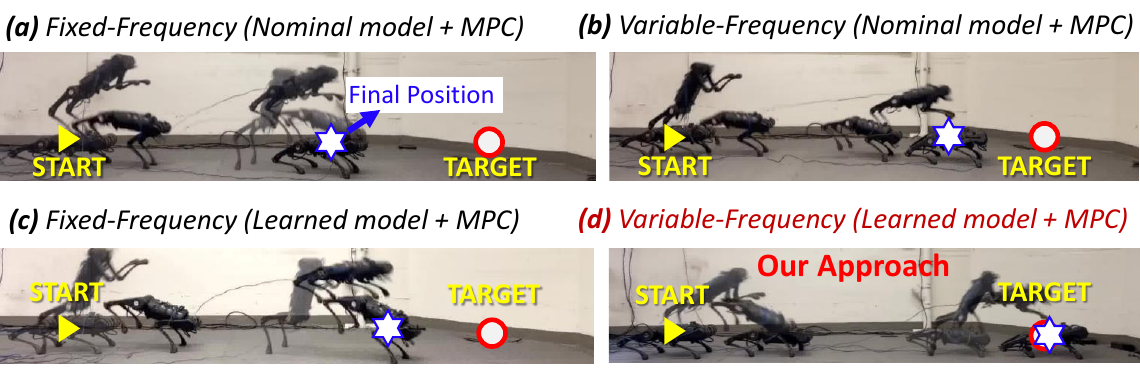}
    \caption{\textbf{Hardware experiments:} \rebut{We conduct $2$ rounds with $4$ jumps in each round for each method. The figures (a)-(d) select snapshots of the first and the last jumps when the robot executes four consecutive jumps with different MPC approaches. The target for each jump is $0.6$ m.}} \label{fig:compare_all_mpcs_hardware}
    \includegraphics[width=\linewidth,trim={3.5cm 9.2cm 3.7cm 9cm},clip]{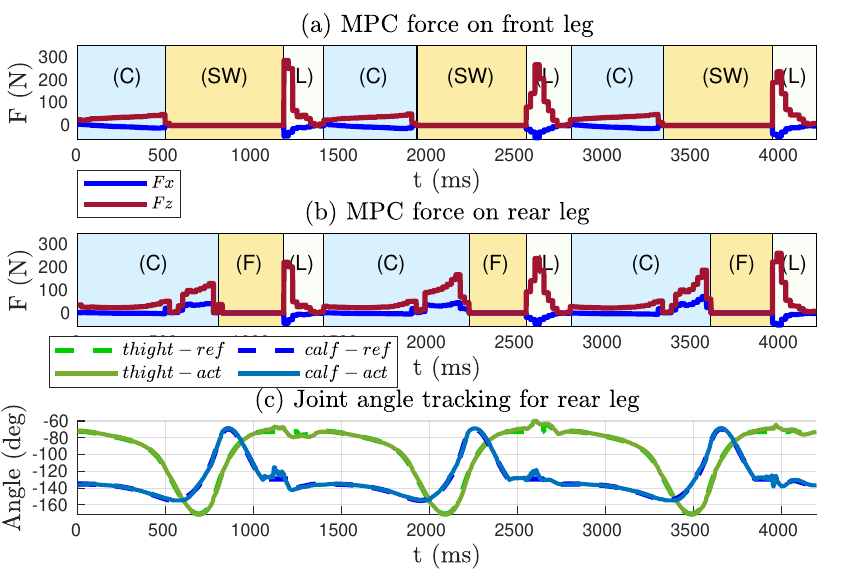}
    \caption{\textbf{Continuous jumping on uneven terrain with A1 robot, shown in Fig. \ref{fig:introduction}:} control forces on the front (a) and rear (b) legs, where the forces are $0$ for swing legs;
    and (c) accurate joint tracking of thigh and calf of rear legs. \rebut{Plots (a) and (b) show the force outputs satisfy $|F_z| \leq F_{max} = 350$ N and the friction cone $|F_x/F_z| \leq \mu = 0.6$ during all phases.}}
    \label{fig:uneven-terrain-mpc-hardware}
\vspace{-0.5cm} 
\end{figure}

\redst{For further evaluation of target accuracy and robustness,  we performed 4 consecutive jumps (targeting $0.6$ m per jump), repeated 2 times, on flat ground for each of four MPC setups} 
We \redst{further}evaluate the target accuracy and robustness with our method for continuous jumps on flat ground.
The results are presented in Fig. \ref{fig:compare_all_mpcs_hardware}, showing some selected snapshots of the first and the last jump to the final target $2.4$ m. Our method achieves the highest target accuracy, allowing the robot to traverse $2.47$ m with an average distance error of only $1.75$ cm per jump.
During flight periods, the robot can jump up to $4\times$ its standing height.



Note that computing residual matrices $\bfG_{\calC,\bftheta}, \bfH_{\calC,\bftheta}, \bfH_{\calF,\bftheta}$ can consume significant time if we feed the neural network with $(\bfx_0, \bfu_0), (\bfx_{1}^*, \bfu_{1}^*), ..., (\bfx_{K}^*, \bfu_{K}^*)$ sequentially. To improve computational efficiency, we combined the current state and reference as a batch, then feed-forward the entire batch to the learned neural networks all at once, reducing the neural network query time to less than $1$ ms.

\rebut{
We further evaluate MPC with different models learned with other timing choices of $(\Delta t_c, \Delta t_f)= \lbrace (50,50), (25, 80)\rbrace$ ms, in addition to the two timesteps $(\Delta t_c, \Delta t_f)= \lbrace (25, 25), (25, 100) \rbrace$ we already discussed above.
}
\begin{table}[t]
	\centering
	\caption{Distance errors (cm) when varying coarseness of timesteps $(\Delta t _{\calC}, \Delta t _{\calF})$ ms. Red colors indicate jumping failure as robot falls short of the box (positive values). The distance errors are measured by the motion capture system at the end of the predefined flight period ($1200$ms). Supplemental video: \url{https://youtu.be/2QjZkARs1mU}.}
	\begin{tabular}{ccccc}
		\hline
		Task & $(25,25)$ & $(50,50)$ & $(25,80)$ & $(25,100)$ \\[.5ex]
		\hline
        Box-Jump & \textcolor{red}{$18.3 \pm 3.7$} & \textcolor{red}{$10.8 \pm 2.1$} & $-1.8 \pm 0.9$ & $-2.0 \pm 1.3$ \\[.5ex]
		\hline 
		\label{tab:vary_timestep}
	\end{tabular}
  \vspace{-3.6em}
\end{table}
\rebut{Three jumps are conducted separately for each timing choice of jumping on box task, and the experiments are summarized in Tab. \ref{tab:vary_timestep}. We also achieved successful jumps using the variable-frequency scheme $(25, 80)$ ms. We also observed that using higher timestep in contact to cover the whole flight phase even with fixed-frequency scheme, e.g.,$( 50, 50)$ ms does not guarantee successful jumps.
This confirms our rationale that due to the higher complexity of the robot dynamics during the contact phase, a
coarse time step during this phase leads to higher accumulated prediction errors, even in the flight phase, causing task
failures.

}

\subsection{Continuous Jumping on Uneven Terrain}
We tested the robustness and target accuracy of our learning-based MPC for continuous jumping on uneven terrains with a target of $60$ cm for each jump, as illustrated in Fig. \ref{fig:introduction} and Fig. \ref{fig:uneven-terrain-mpc-hardware}. The terrain consisted of multiple blocks with random heights between $2-4$ cm, randomly placed on the ground. 
\rebut{Since the robot legs can impact the ground early or late on the real robot, there is usually a mismatch between scheduled contact and actual contact states. Whenever both contacts happen (i.e., the landing phase $L$ starts), we activate a separate landing MPC to make a transition between two jumps in a short period of $\sim$$200$ ms \cite{continuous_jump}. The landing MPC is designed with simplified dynamics, $K=10$, and $\Delta t=25$ ms. This aims to track the body reference trajectory from TO for continuous jumping \cite{continuous_jump} and connects two jumps seamlessly.}
\redst{Fig. \ref{fig:uneven-terrain-mpc-hardware}(a)}\rebut{Figure \ref{fig:introduction}} shows the actual trajectory of three continuous jumps, leaping around $175$ cm in total that yields only $1.67$ cm distance error ($<3\%$) for each jump on the uneven terrain. \rebut{Compared to our prior work \cite{continuous_jump}, we successfully achieve continuous jumping in hardware in this work.}
\redst{Fig. \ref{fig:uneven-terrain-mpc-hardware}(b) and (c) show the force outputs \redst{from adaptive-frequency MPC for the learned model. The force controls} on both legs satisfy $|F_z| \leq F_{max} = 350$ N and the friction cone $|F_x/F_z| \leq \mu = 0.6$ during all phases.} 
With our framework, we also achieve a target jumping error of less than $2$ cm ($3.5\%$) on $34$ single and continuous jumps with different targets and uneven terrain.

\section{Discussion and Conclusion} \label{sec:conclusion}
In this letter, we developed a learned-model MPC approach that enables both target accuracy and robustness for aggressive \rebut{planar} jumping motions. We learned a residual model from a limited dataset to account for the effect of leg dynamics and model mismatch with real hardware. Given the learned model, we designed \rebut{a learning-based variable-frequency MPC that considers the jumping transitions, the entire flight phase, and the jumping target} during the optimization process. 
\rebut{We demonstrated the scalability of our approach to handle new jumping targets under unseen reference trajectories or unseen jumping tasks. Linearized MPC sacrifices a certain degree of accuracy if the system does not operate around the reference under highly nonlinear conditions or substantial disturbances. Our approach introduces a learned residual model to compensate for linearization errors and disturbances without sacrificing the real-time computational efficiency. Our future work will explore nonlinear MPC with learned dynamics and its higher computational cost. We will also incorporate line-foot, rolling, and soft contact in learning aggressive legged locomotion maneuvers.}

\bibliographystyle{ieeetr}
\bibliography{reference}

\end{document}